\documentclass[runningheads]{llncs}
\usepackage{graphicx}
\usepackage{color}  
\usepackage{enumitem}
\usepackage{amsmath}
\usepackage{amssymb}
\usepackage{booktabs} 
\usepackage{multirow}
\usepackage{float}

\usepackage{hyperref} 
\definecolor{myblue}{RGB}{60,120,216}
\hypersetup{
  colorlinks=true,
  linkcolor=myblue,
  citecolor=myblue,
  urlcolor=myblue
}

\setlist[itemize]{label=\textbullet}

\begin{document}
\title{
KSG-Net: Key-Sparse and Global-Context Learning for Maritime 3D Ship Detection
}
%



\author{Zhouyuan Huai\inst{1} \and
Meiqi Wan\inst{1} \and
Yan Yang\inst{3,4} \and 
Minshi Chen\inst{1} \and 
Xin Yuan\inst{1,2,}$^{*}$\\
Wei Wang\inst{1,2} \and
Xiao Wang\inst{1,2}}
\institute{School of Computer Science and Technology, Wuhan University of Science and Technology, Wuhan 430065, China \and
Hubei Province Key Laboratory of Intelligent Information Processing and Real-Time Industrial System, Wuhan University of Science and Technology, Wuhan 430065, China \and
State Key Laboratory of Robotics and Intelligent Systems, Shenyang Institute of Automation, Chinese Academy of Sciences, Shenyang 110016, China \and
China University of Chinese Academy of Sciences, Beijing 100049, China}

\renewcommand{\thefootnote}{}
\footnotetext{$^{*}$ Corresponding author: Xin Yuan (yuanxincherry@gmail.com)}
\maketitle              

\begin{abstract}
Accurate 3D ship detection in maritime environments is critical for autonomous navigation, yet remains challenging due to large-scale vessel variations, sparse point clouds of small vessels, and severe sea-clutter interference. Existing methods, primarily based on 2D features or dense repre sentations, struggle to balance detection accuracy and computational efficiency, while sparse 3D detectors designed for road scenes generalize poorly to maritime scenarios. This paper focuses on two key challenges in maritime LiDAR perception: weak feature representation for small and sparse vessels, and insufficient global structural modeling for large vessels due to the limited receptive field of local sparse convolutions. To address these issues, we propose KSG-Net, a Key-Sparse and Global-Context learning network for maritime 3D ship detection. The core idea is to jointly enhance local discriminative features and global structural awareness within a unified fully sparse detection framework. Specifically, a Key Sparse Multi-scale Aggregation (KSMA) module is designed to enhance the representation of small and sparse vessels by selecting informative key voxels and aggregating cross-scale neighborhood features. Furthermore, a Global Context Aggregation (GCA) module is introduced to capture long-range geometric dependencies through scene-level context modeling with gated residual interactions, thereby improving the representation of large vessels. Extensive experiments on the Thames River vessel dataset and simulated datasets demonstrate that KSG-Net consistently outperforms existing methods in multi-scale vessel detection and exhibits strong robustness in complex maritime environments.

\end{abstract}

\section{Introduction}

Accurate 3D ship detection in maritime environments is essential for autonomous navigation, collision avoidance, and intelligent decision-making. Compared with road scenes, maritime environments exhibit more complex sensing conditions, including large vessel-scale variations, irregular background structures, sparse point observations, and severe sea-clutter interference~\cite{lin2022maritime,xie2024reliable}. These characteristics make maritime 3D ship detection substantially different from conventional autonomous-driving detection tasks.


Existing maritime perception methods are still largely dominated by image-based detection, dense 3D representations, or multi-sensor fusion pipelines~\cite{wang2025hierarchical,xie2026uncertainty}. Although these methods have achieved meaningful progress, they often suffer from quantization errors and computational redundancy during the densification process, limiting their applicability in large-scale maritime environments. Meanwhile, fully sparse 3D detectors developed for road scenes, such as VoxelNeXt~\cite{chen2023voxelnext}, provide an efficient paradigm by operating directly on non-empty sparse voxels. However, directly transferring these frameworks to maritime scenarios is non-trivial, as maritime targets exhibit more severe scale variation and less regular point distributions.

\begin{figure}[htbp]
\centering
\includegraphics[width=\linewidth]{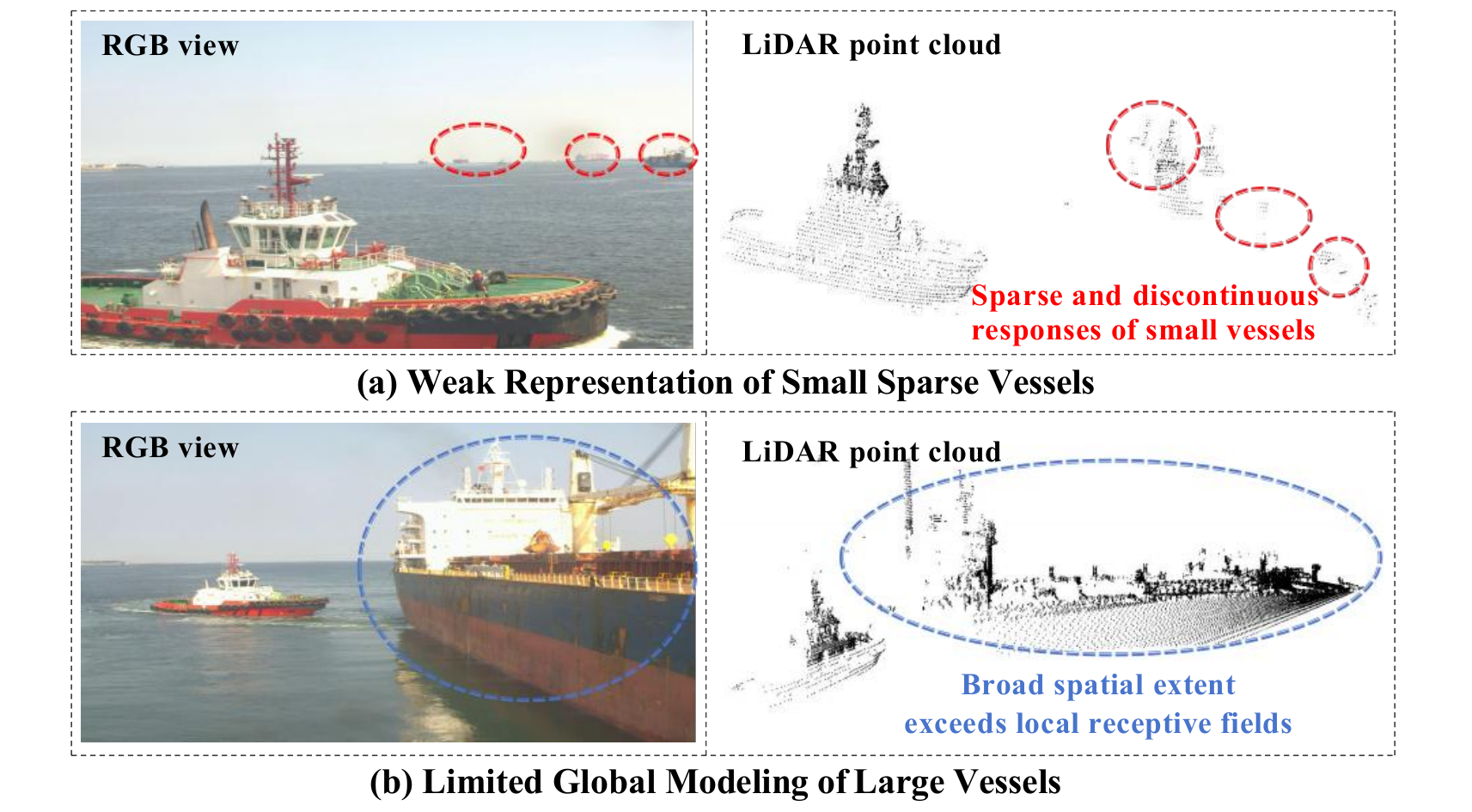}
\vspace{-6mm}
\caption{
Illustration of two key challenges in maritime 3D ship detection.
\textbf{(a) Weak Representation of Small Sparse Vessels}. Small vessels usually produce sparse and discontinuous point responses, leading to weak and unstable local representations.
\textbf{(b) Limited Global Modeling of Large Vessels}. Large vessels often span broad spatial regions, while local sparse convolutions have limited receptive fields and struggle to capture long-range structural dependencies.}
\label{fig:problem_overview}
\end{figure}


As shown in Fig.~\ref{fig:problem_overview}, two key challenges arise in maritime 3D ship detection: (1) Weak representation of small and sparse vessels. Small vessels often produce only sparse and discontinuous point responses, providing limited geometric cues that are easily confused with background clutter. (2) Insufficient global structural modeling for large vessels. Large vessels usually span broad spatial regions, while local sparse convolutions have limited receptive fields and struggle to capture long-range structural dependencies. These scale-dependent challenges motivate the need for a unified sparse detection framework that can simultaneously enhance local discriminative features for small vessels and strengthen global structural modeling for large vessels.


To address the above challenges, we propose KSG-Net, a Key-Sparse and Global-Context Detection Network tailored for maritime 3D ship detection. The core idea is to jointly enhance local discriminative representations and global structural awareness within a unified sparse detection framework. To this end, we design two complementary modules: (1) a Key Sparse Multi-scale Aggregation (KSMA) module, which enhances the representation of small and sparse vessels by selecting informative key voxels and performing cross-scale neighborhood aggregation under highly sparse point distributions; and (2) a Global Context Aggregation (GCA) module, which captures long-range geometric dependencies by modeling scene-level global context and injecting it into sparse features through gated residual interactions, thereby improving the representation of large vessels and robustness in cluttered maritime environments.
In summary, the main contributions of this paper are as follows:
\begin{itemize}
    \item 
    We propose KSG-Net, a maritime-oriented fully sparse 3D detection framework for vessel detection in LiDAR point clouds, addressing the overlooked challenges of weak small-vessel representation and insufficient large-vessel global modeling.
    \item 
    We design the Key Sparse Multi-scale Aggregation (KSMA) module, which enhances the representation of small and sparse vessels through informative key-voxel selection and cross-scale neighborhood interaction, focusing computation on informative sparse locations.
    \item 
    We introduce the Global Context Aggregation (GCA) module, which addresses the limited receptive field of local sparse convolutions by capturing long-range geometric structures for large vessels through scene-level global context modeling and gated residual injection.
    \item 
     Extensive experiments on the Thames River vessel dataset and simulated datasets demonstrate that KSG-Net outperforms existing methods in multi-scale vessel detection and exhibits strong robustness in complex maritime environments.
\end{itemize}

\section{Related Work}

\subsection{LiDAR-Based Dense Detectors}
Dense detectors convert irregular point clouds into structured representations, such as bird’s-eye view (BEV) feature maps, enabling the use of mature convolutional neural networks. Early works, including MV3D and PointPillars~\cite{lang2019pointpillars}, establish the paradigm of projecting point clouds into dense representations for efficient processing, while CenterPoint~\cite{yin2021center} further improves localization accuracy via center-based object modeling.
Recent dense LiDAR detectors have further improved representation capability through architectural refinement, temporal modeling, and multi-modal fusion. L4DR~\cite{huang2025l4dr} proposes a LiDAR and 4D radar fusion framework that improves detection robustness under adverse weather conditions. BevNext~\cite{li2024bevnext} modernizes dense BEV-based detection pipelines by integrating CRF-based depth estimation and long-term temporal feature aggregation, achieving strong performance on the nuScenes benchmark. FSD-BEV~\cite{jiang2024fsd} introduces foreground self-distillation and frame-level point cloud densification strategies, improving single-frame BEV detection performance.

Despite their effectiveness, dense detectors inevitably introduce quantization errors and computational redundancy during the densification process, which becomes more pronounced in large-scale maritime environments. Moreover, these methods often struggle to preserve fine-grained geometric structures of small and sparse vessels, limiting their applicability in maritime LiDAR scenarios.

\subsection{LiDAR-Based Sparse Detectors}
Sparse 3D detectors improve the efficiency of LiDAR-based 3D object detection by performing feature extraction and prediction only on non-empty voxels or foreground points, avoiding the redundancy introduced by dense feature construction. Representative methods such as SECOND~\cite{yan2018second}, PV-RCNN~\cite{shi2020pv}, and PV-RCNN++~\cite{shi2023pv} have advanced this direction through sparse voxel encoding and point-voxel collaborative representation learning.

Recent studies have further developed fully sparse detection frameworks. FSD~\cite{fan2022fully} systematically constructs a fully sparse 3D detector, VoxelNeXt~\cite{chen2023voxelnext} unifies proposal generation and detection heads in sparse voxel space, and FSD V2~\cite{fan2024fsd} improves instance-level representation with virtual voxels. In parallel, long-range dependency modeling has also received increasing attention. HEDNet~\cite{zhang2023hednet}, SAFDNet~\cite{zhang2024safdnet}, DSVT~\cite{wang2023dsvt}, and LION~\cite{liu2024lion} enhance sparse representations through hierarchical interaction, adaptive feature diffusion, sparse Transformer modeling, and state-space-style feature interaction, respectively.

Although sparse detectors have achieved remarkable progress in efficiency and long-range perception, most of them are designed for autonomous driving scenarios, where object layouts are relatively structured and backgrounds are more controlled. Maritime scenes present more irregular target distributions, stronger background interference, and larger scale variations. Small vessels often produce extremely sparse point responses, while large vessels span broad spatial regions. Therefore, existing sparse detectors remain limited in stable small-target representation and global structural modeling for large vessels. Motivated by these limitations, our work builds upon VoxelNeXt~\cite{chen2023voxelnext} and further enhances key sparse representation and global context modeling for maritime LiDAR 3D detection.

\subsection{LiDAR-based Maritime Perception}
Maritime perception has attracted increasing attention in recent years, and LiDAR has shown strong potential for object detection, tracking, and environment understanding in maritime environments. Lin \textit{et al.}~\cite{lin2022maritime} provided a systematic review of deep-learning-based maritime perception and highlighted the substantial differences between maritime and road scenes in terms of target scale, background structure, and environmental disturbances. Xie \textit{et al.}~\cite{xie2024reliable} further advanced LiDAR-based ship detection and tracking in busy maritime environments, while recent datasets have provided important support for maritime LiDAR perception research~\cite{martelli2026arnold,zhang2024lidar}.

Despite these advances, many maritime perception methods still rely on image-based detection, multi-sensor fusion, dense 3D representations, or non-fully-sparse point-cloud detection paradigms~\cite{lang2019pointpillars,lin2022maritime,shi2020pv}. Recent maritime 3D detection methods have begun to address vessel scale variation and complex marine conditions through hierarchical vertical-aware modeling, adaptive multi-scale design, and uncertainty-aware point cloud detection~\cite{wang2025hierarchical,xie2026uncertainty}. However, a unified fully sparse framework tailored to maritime LiDAR detection remains underexplored, especially for jointly handling stable representation of small sparse vessels and global structural modeling of large vessels. Different from existing maritime detectors, our method starts from a fully sparse framework and designs KSMA and GCA to enhance key sparse representation and global context modeling, respectively.
\section{Methodology}
\begin{figure}[!t]
\centering
\includegraphics[
width=0.98\textwidth
]{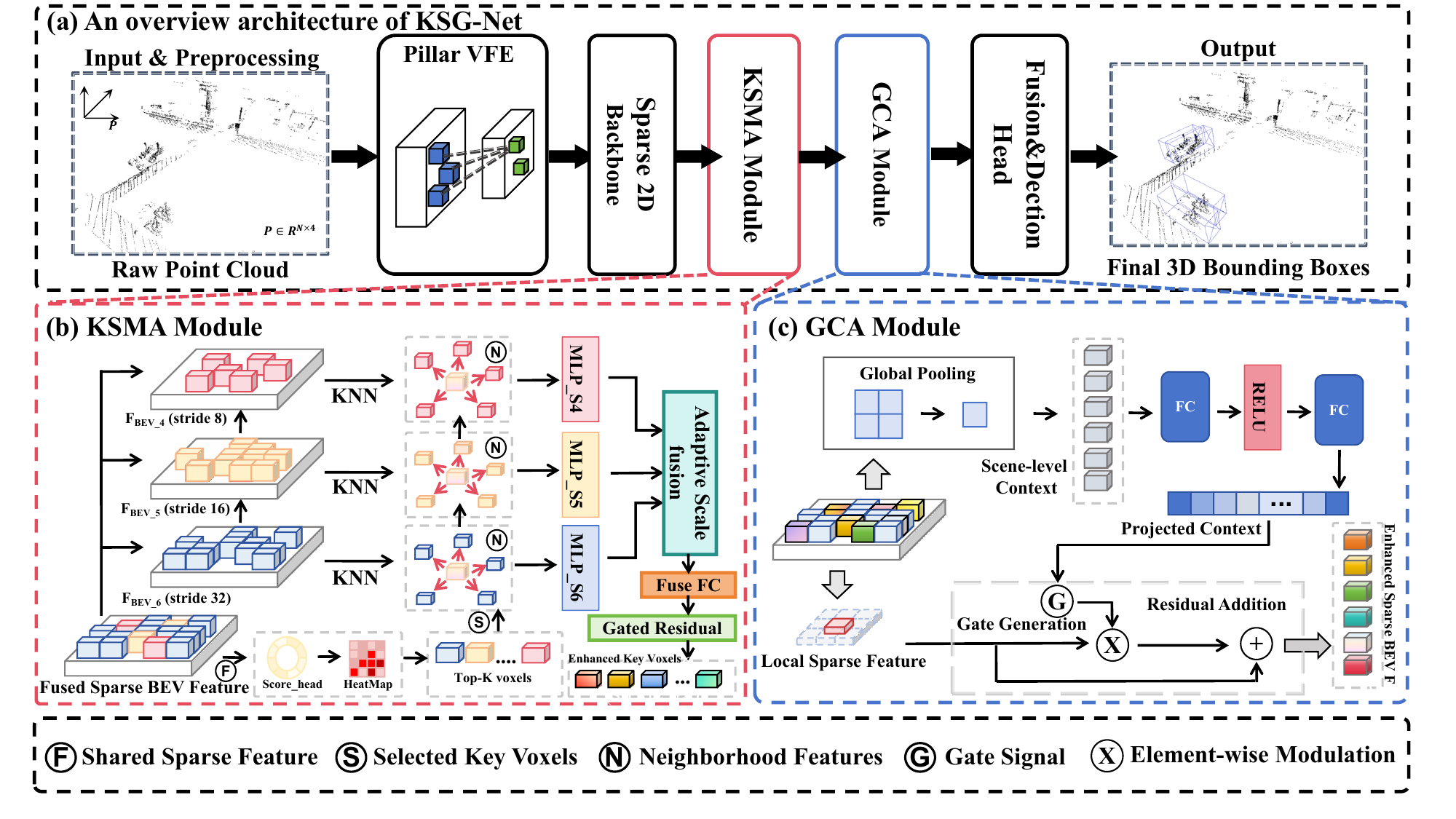}
\vspace{-3mm}
\caption{Overview of KSG-Net. (a) The overall fully sparse detection pipeline for maritime LiDAR point clouds. (b) The Key Sparse Multi-scale Aggregation (KSMA) module, which enhances informative sparse locations through key-voxel identification and cross-scale neighborhood aggregation. (c) The Global Context Aggregation (GCA) module, which injects scene-level global context into the sparse representation to improve large-vessel structural modeling.}
\label{fig:framework}
\vspace{-3mm}
\end{figure}
In this section, we present the proposed KSG-Net. Fig.~\ref{fig:framework} illustrates the overall architecture of KSG-Net and its two core modules. As shown in Fig.~\ref{fig:framework}(a), KSG-Net follows a fully sparse 3D detection paradigm, consisting of input encoding, sparse backbone feature extraction, KSMA, GCA, and the detection head. Fig.~\ref{fig:framework}(b) and Fig.~\ref{fig:framework}(c) further show the internal structures of the KSMA and GCA modules, respectively.

Specifically, the input maritime LiDAR point cloud is first encoded by the pillar encoder and processed by the sparse backbone to obtain a shared sparse BEV representation. Based on this representation, KSMA enhances local discriminative features for small sparse vessels through key-voxel identification and multi-scale neighborhood aggregation. GCA then supplements long-range structural information for large vessels through scene-level global context modeling and gated residual injection. The two modules are sequentially applied to the same shared sparse representation, improving multi-scale vessel detection while preserving the efficiency of fully sparse inference.

\subsection{Key Sparse Multi-scale Aggregation}
The objective of KSMA is to selectively enhance local discriminative features for small sparse vessels. As shown in Fig.~\ref{fig:framework}(b), KSMA first identifies informative key voxels from the shared sparse representation, aggregates their multi-scale sparse neighborhoods, and writes the refined responses back through adaptive scale fusion and residual enhancement. This \emph{selecting-before-enhancing} strategy focuses computation on informative sparse locations and improves local representation under sparse maritime observations.

\subsubsection{Key Sparse Voxel Identification}

Let the shared sparse BEV representation be denoted as $\mathcal{F}=\{(\mathbf{u}_i,\mathbf{x}_i)\}_{i=1}^{M}$, where $\mathbf{u}_i \in \mathbb{Z}^{2}$ denotes the BEV coordinate of the $i$-th non-empty voxel and $\mathbf{x}_i \in \mathbb{R}^{C}$ denotes the corresponding feature vector. To estimate the importance of each sparse location, we introduce a lightweight sparse scoring head on top of the shared representation and produce a saliency score for every voxel:
\begin{equation}
\omega_i = \sigma\!\left(\Psi_{\mathrm{score}}(\mathbf{x}_i)\right),
\label{eq:ksma_score}
\end{equation}
where $\Psi_{\mathrm{score}}(\cdot)$ denotes the sparse scoring function and $\sigma(\cdot)$ is the sigmoid activation.

Based on these scores, we retain the top-$K$ voxels within each sample and treat them as the key-voxel set. Let $\mathcal{I}_b$ denote the voxel index set of the $b$-th sample. The corresponding key-voxel set is defined as
\begin{equation}
\mathcal{K}_b =
\operatorname{TopK}
\left(
\{\omega_i \mid i \in \mathcal{I}_b\}, K
\right),
\qquad
\mathcal{K} = \bigcup_{b=1}^{B}\mathcal{K}_b.
\label{eq:ksma_topk}
\end{equation}
This saliency-aware selection avoids uniformly enhancing all non-empty voxels and instead uses high-response locations as anchors for subsequent aggregation. It therefore concentrates computation on vessel-related sparse structures and reduces interference from background clutter.

\subsubsection{Multi-scale Sparse Aggregation and Feature Refinement}

After key voxel selection, KSMA aggregates neighborhood context from multi-scale sparse BEV features. Let $\mathcal{F}^{(l)}={(\mathbf{u}^{(l)}_j,\mathbf{x}^{(l)}*j)}*{j=1}^{M_l}$ denote the sparse feature set at scale $l$. For each key voxel $i \in \mathcal{K}$, its coordinate is aligned to scale $l$ by $\delta_l$, followed by $k$-nearest-neighbor search:
\begin{equation}
\tilde{\mathbf{u}}_i^{(l)} = \frac{\mathbf{u}_i}{\delta_l},
\qquad
\mathcal{N}_i^{(l)} =
\operatorname{KNN}
\left(
\tilde{\mathbf{u}}_i^{(l)}, \{\mathbf{u}^{(l)}_j\}_{j=1}^{M_l}, k_l
\right),
\label{eq:ksma_knn}
\end{equation}
where $\delta_l$ and $k_l$ denote the scale alignment factor and neighborhood size at scale $l$, respectively.

For each scale, the neighboring features are first mean-aggregated and then projected through a scale-specific transformation function to obtain a scale-wise aggregated representation:
\begin{equation}
\bar{\mathbf{z}}_i^{(l)} =
\frac{1}{|\mathcal{N}_i^{(l)}|}
\sum_{j \in \mathcal{N}_i^{(l)}} \mathbf{x}^{(l)}_j,
\qquad
\mathbf{z}_i^{(l)} = \Phi_l\!\left(\bar{\mathbf{z}}_i^{(l)}\right),
\label{eq:ksma_scale_agg}
\end{equation}
where $\Phi_l(\cdot)$ denotes the projection function associated with scale $l$.

Since the contributions of different scales may vary for different key voxels, we further predict adaptive scale weights from the current key feature and fuse the scale-wise aggregated representations accordingly:
\begin{equation}
\boldsymbol{\alpha}_i =
\operatorname{softmax}(\mathbf{W}_{\alpha}\mathbf{x}_i),
\qquad
\tilde{\mathbf{z}}_i =
\sum_{l=1}^{L}\alpha_i^{(l)}\mathbf{z}_i^{(l)},
\label{eq:ksma_fusion}
\end{equation}
where $\boldsymbol{\alpha}_i = [\alpha_i^{(1)},\dots,\alpha_i^{(L)}]$ denotes the normalized scale-weight vector.

Based on the fused multi-scale context, we concatenate the original key feature with the aggregated feature and transform them into an enhanced representation. Furthermore, since different key voxels may have different confidence levels, we use the saliency score as a gating factor to modulate the residual update magnitude, yielding the final KSMA output:
\begin{equation}
\hat{\mathbf{x}}_i =
\Phi_{\mathrm{fuse}}
\left(
[\mathbf{x}_i \,\|\, \tilde{\mathbf{z}}_i]
\right),
\label{eq:ksma_refine}
\end{equation}
\begin{equation}
\mathbf{x}_i^{\mathrm{KSMA}} =
\mathbf{x}_i + \gamma \, \omega_i \, \hat{\mathbf{x}}_i,
\qquad i \in \mathcal{K},
\label{eq:ksma_update}
\end{equation}
where $\Phi_{\mathrm{fuse}}(\cdot)$ denotes the feature fusion mapping and $\gamma$ is a learnable residual scaling coefficient.

In this way, KSMA enhances only informative sparse locations with cross-scale neighborhood context, improving small-vessel representations while preserving the efficiency of fully sparse inference.
\subsection{Global Context Aggregation}

Although KSMA significantly improves the local representation quality of key sparse locations, large vessels in maritime scenes usually exhibit broader spatial extent and more complex geometry. Relying solely on local sparse convolutions and local enhancement remains insufficient to capture their long-range structural dependencies. To address this issue, we further design GCA to inject scene-level global context into the sparse representation in a lightweight manner, thereby improving the holistic understanding of large-scale structures. As illustrated in Fig.~\ref{fig:framework}(c), GCA mainly consists of scene-level global context modeling and context-gated residual enhancement.

\subsubsection{Scene-level Global Context Modeling}

Let the sparse feature set after KSMA be denoted as $\mathcal{F}^{\mathrm{KSMA}}=\{(\mathbf{u}_i,\mathbf{x}_i^{\mathrm{KSMA}})\}_{i=1}^{M}$. For each sample, we first perform global pooling over all non-empty sparse locations to extract a scene-level context descriptor, and then project it into the same representation space as the local sparse features:
\begin{equation}
\mathbf{g}_b =
\rho
\left(
\{\mathbf{x}_i^{\mathrm{KSMA}} \mid i \in \mathcal{I}_b\}
\right),
\label{eq:gca_pool}
\end{equation}
\begin{equation}
\bar{\mathbf{g}}_b = \Phi_{\mathrm{ctx}}(\mathbf{g}_b),
\label{eq:gca_proj}
\end{equation}
where $\rho(\cdot)$ denotes a global pooling operator and $\Phi_{\mathrm{ctx}}(\cdot)$ is the context projection mapping.   

This process compresses scene statistics originally distributed across different sparse local positions into a unified global descriptor, which serves as a scene-level reference for the subsequent enhancement of large-scale structures. Unlike global self-attention that explicitly models pairwise interactions among all voxels, this pooling-based context modeling is considerably lighter and better suited to the efficiency requirements of fully sparse detection.

\subsubsection{Context-Gated Residual Enhancement}

After obtaining the scene-level context, we broadcast it back to each non-empty voxel in the current sample and use a gating mechanism to adaptively regulate the context injection strength. For the $i$-th voxel in sample $b$, the gate vector is jointly predicted from the local feature and the global context:
\begin{equation}
\mathbf{m}_i =
\sigma
\left(
\Phi_{\mathrm{gate}}
\left(
[\mathbf{x}_i^{\mathrm{KSMA}} \,\|\, \bar{\mathbf{g}}_b]
\right)
\right),
\label{eq:gca_gate}
\end{equation}
where $\Phi_{\mathrm{gate}}(\cdot)$ denotes the gating function.

The global context is then injected into the local feature in a channel-wise gated manner, and its overall influence is controlled by a learnable residual coefficient, yielding the final GCA-enhanced representation:
\begin{equation}
\mathbf{x}_i^{\mathrm{GCA}} =
\mathbf{x}_i^{\mathrm{KSMA}}
+
\lambda
\left(
\mathbf{m}_i \odot \bar{\mathbf{g}}_b
\right),
\label{eq:gca_update}   
\end{equation}
where $\odot$ denotes element-wise multiplication and $\lambda$ is a learnable residual scaling factor.

This gated residual injection allows the model to adaptively determine how much global information should be injected according to the local feature content at each sparse position. For large vessels in maritime scenes, this design effectively alleviates the fragmented representation caused by limited local receptive fields and enhances the modeling of long-range geometric dependencies.

After KSMA and GCA enhancement, the refined sparse features are fed into the detection head for final prediction. Let the prediction set be denoted as $\mathcal{Y}$, which includes the classification heatmap, center offsets, vertical offset, box dimensions, orientation, and an IoU-related branch. During training, we follow the objective design of the fully sparse baseline detector, and the overall loss is formulated as
\begin{equation}
\mathcal{Y} =
\left\{
\mathbf{H},
\Delta \mathbf{c},
\Delta z,
\mathbf{d},
\mathbf{r},
q
\right\},
\label{eq:head_output}
\end{equation}
\begin{equation}
\mathcal{L} =
\mathcal{L}_{\mathrm{hm}}
+
\lambda_{\mathrm{reg}}\mathcal{L}_{\mathrm{reg}}
+
\lambda_{\mathrm{iou}}\mathcal{L}_{\mathrm{iou}}
+
\lambda_{\mathrm{iou\_reg}}\mathcal{L}_{\mathrm{iou\_reg}},
\label{eq:total_loss}
\end{equation}
where $\mathbf{H}$ denotes the sparse classification heatmap, $\Delta \mathbf{c}$ denotes the 2D center offset on the BEV plane, $\Delta z$ denotes the vertical offset, $\mathbf{d}$ and $\mathbf{r}$ denote box dimensions and orientation, and $q$ denotes the IoU prediction. With this formulation, KSG-Net can be trained end-to-end within a unified fully sparse detection framework.
\section{Experiments}


\subsection{Experimental Setup}
\textbf{Ship LiDAR Dataset: } We use the maritime LiDAR point cloud dataset publicly released by Zhang \textit{et al.}~\cite{zhang2024lidar}. The dataset was collected using a multi-beam photon-counting LiDAR sensor and contains real-world port scenes from Wuhan, Shanghai, and Qingdao. The annotated objects are categorized into four vessel classes: \textit{Cargo ship}, \textit{Tour boat}, \textit{Engineering ship}, and \textit{Speedboat}. The dataset is organized in KITTI format and consists of 10,571 training frames, 1,200 validation frames, and 1,500 testing frames.

\noindent \textbf{Thames River Real Dataset: } We use the first real-world Thames River LiDAR dataset released by Xie \textit{et al.}~\cite{xie2024reliable}. The dataset was collected in the London docklands and busy Thames waterways, covering diverse vessel types and scales, including monohulls, catamarans, and sailboats with lengths ranging from approximately 5~m to 40~m. It contains 9,200 frames in total, with 7,820 frames for training and 1,380 frames for testing, corresponding to an 85\%/15\% split.

\noindent \textbf{Thames River Simulated Dataset: } This dataset consists of synthetic data generated using the Robot Operating System (ROS), simulating port and fairway environments with maritime targets such as ships and buoys. The raw simulated data were further refined by Xie \textit{et al.}~\cite{xie2024reliable} to correct annotation errors and improve data quality. The version used in this study contains 11,360 frames, covering diverse scenarios with isolated vessel point clouds as well as complex terrain backgrounds.

\subsubsection{Experimental Details}
We evaluate all methods using KITTI-style 3D Average Precision (AP) and mean Average Precision (mAP) under the $R_{40}$ protocol. A detection is considered correct when the predicted class matches the ground truth and the 3D IoU between the predicted and ground-truth boxes is greater than 0.5. We therefore report 3D AP and 3D mAP at an IoU threshold of 0.5 for all datasets.
All experiments are implemented with OpenPCDet and PyTorch on a workstation equipped with an Intel Core i9-14900KF CPU and a single NVIDIA GeForce RTX 4060 GPU. The compared detectors were originally developed for terrestrial autonomous-driving scenes. To ensure a fair comparison, we retrain all methods on the same maritime data splits with the same point-cloud range, input preprocessing, and evaluation protocol. For each baseline, we follow its official training configuration as closely as possible, including the optimizer and learning-rate schedule, and adapt only the class definitions and detection range to the maritime setting. KSG-Net is trained with the Adam optimizer and a OneCycle learning-rate scheduler, using a peak learning rate of $1\times10^{-3}$, a weight decay of 0.01, a batch size of 4, and 80 epochs.

\begin{table*}[htbp]
    \centering
    \tiny
    \setlength{\tabcolsep}{3.5pt}
    \renewcommand{\arraystretch}{1.10}
    \caption{Quantitative comparison on the Ship LiDAR Dataset. AP@0.5 (\%) is reported for representative categories, where \textit{Cargo}, \textit{Tour}, \textit{Engineer}, and \textit{Speed} denote cargo ships, tour boats, engineering ships, and speedboats, respectively. FLOPs are effective FLOPs averaged over validation samples. FPS is measured with a batch size of 1. GBlobs is evaluated as a local geometric encoder on VoxelNeXt. The best AP results are highlighted in \textbf{bold}. ``--'' denotes a result that is not yet available.}
    \label{tab:ship_lidar_results}
    \resizebox{\linewidth}{!}{
    \begin{tabular}{c|l|cccc|cc|c}
        \toprule
        & \multirow{2}{*}{Method} & \multicolumn{4}{c|}{AP@0.5 (\%)} & \multirow{2}{*}{FLOPs (G)} & \multirow{2}{*}{FPS (Hz)} & \multirow{2}{*}{\textbf{mAP} (\%)} \\
        \cmidrule(lr){3-6}
        & & Cargo & Tour & Engineer & Speed & & & \\
        \midrule
        \multirow{3}{*}{\rotatebox{90}{Point}}
        & PointRCNN (CVPR 2019) \cite{shi2019pointrcnn} & 64.71 & 76.16 & 78.76 & 30.00 & 127.41& 10.37 & 62.40 \\
        & Part-A2 (CVPR 2020) \cite{shi2019part}      & 63.89 & 78.33 & 71.74 & 33.78 & 508.35 & 13.19 & 61.94 \\
        & PV-RCNN (CVPR 2020) \cite{shi2020pv}          & 65.81 & 79.30 & 74.21 & 31.02 & 89.02 & 8.89 & 63.09 \\
        \midrule
        \multirow{6}{*}{\rotatebox{90}{Voxel}}
        & PointPillars (CVPR 2019) \cite{lang2019pointpillars} & 57.08 & 76.50 & 85.98 & 42.86 & 63.44 & 62.15 & 65.59 \\
        & VoxelR-CNN (AAAI 2021) \cite{deng2021voxel}         & 60.23 & 77.12 & 78.22 & 41.70 & 48.53 & 25.25 & 64.31 \\
        & VoxelNeXt (CVPR 2023) \cite{chen2023voxelnext}      & 67.65 & 87.81 & 82.49 & 60.07 & 32.03 & 23.55 & 74.50 \\
        & DSVT (CVPR 2023) \cite{wang2023dsvt}                & 71.45 & 87.30 & 82.10 & 68.90 & 339.85 & 14.97 & 77.44 \\
        & FSHNet (CVPR 2025) \cite{liu2025fshnet}             & 67.98 & 80.20 & 84.32 & 70.10 & 54.14 & 12.37 & 75.65 \\
        & GBlobs (CVPR 2025) \cite{malic2025gblobs}          & 65.19 & 78.22 & 87.12 & 63.21 & 32.36 & 22.72 & 74.44 \\
        \midrule
        \multirow{5}{*}{\rotatebox{90}{BEV}}
        & CenterPoint (CVPR 2021) \cite{yin2021center} & 72.16 & 88.52 & 79.60 & 70.22 & 33.23 & 56.65 & 77.63 \\
        & HEDNet (NeurIPS 2023) \cite{zhang2023hednet} & 66.31 & 79.22 & 85.48 & 68.12 & 106.21 & 17.28 & 74.78 \\
        & SAFDNet (CVPR 2024) \cite{zhang2024safdnet}  & 72.31 & 86.98 & 81.24 & 71.75 & 60.83 & 20.25 & 78.07 \\
        & Fade3D (TITS 2025) \cite{ye2025fade3d}       & \textbf{72.90} & 89.50 & 84.15 & 70.45 & 78.31 & 51.53 & 79.25 \\
        & AutoReg3D (CVPR 2026) \cite{huang2026feasibility} & 53.12 & 62.94 & 58.39 & 14.22 & 110.3 & 5.29 & 47.18 \\
        \midrule
        & \textbf{KSG-Net (Ours)} & 72.69 & \textbf{95.15} & \textbf{91.97} & \textbf{82.79} & 31.42 & 33.11 & \textbf{85.65} \\
        \bottomrule
    \end{tabular}}
    \vspace{-6mm}
\end{table*}
\subsection{Comparison on Ship LiDAR Dataset}

The quantitative comparison results on the Ship LiDAR Dataset are summarized in Table~\ref{tab:ship_lidar_results}. KSG-Net achieves the highest mAP@0.5 of $85.65\%$, showing a clear margin over previous methods across point-, voxel-, and BEV-based representations. It surpasses the strongest BEV baseline Fade3D~\cite{ye2025fade3d} by $6.40$\%, SAFDNet~\cite{zhang2024safdnet} by $7.58$\%, and the strongest voxel baseline DSVT~\cite{wang2023dsvt} by $8.21$\%. GBlobs~\cite{malic2025gblobs}, instantiated as a local geometric encoder on VoxelNeXt, reaches $74.44\%$ mAP, close to VoxelNeXt ($74.50\%$) and well below KSG-Net. AutoReg3D~\cite{huang2026feasibility} obtains $47.18\%$ mAP and is the weakest entry, with a particularly low Speedboat AP of $14.22\%$. In terms of efficiency, KSG-Net uses $31.42$~G FLOPs and runs at $33.11$~FPS, remaining comparable to VoxelNeXt and GBlobs while being substantially cheaper than DSVT and AutoReg3D.


A detailed category-wise analysis further supports this conclusion. Fade3D~\cite{ye2025fade3d} leads by only $0.21$ \% on the large and geometrically regular \textit{Cargo} category ($72.90\%$ vs. ours $72.69\%$), whereas KSG-Net achieves the best AP on the more heterogeneous \textit{Tour boat} ($95.15\%$), \textit{Engineering ship} ($91.97\%$), and \textit{Speedboat} ($82.79\%$) categories. The gain on \textit{Speedboat} is $11.04$ \% over SAFDNet~\cite{zhang2024safdnet} ($71.75\%$). Speedboats typically produce the sparsest point responses and the largest shape variation; this improvement is therefore consistent with the design of KSMA for strengthening weak local responses, complemented by GCA for stabilizing structural context.
\subsection{Comparison on Thames River Real and Simulated Datasets}
\begin{table*}[t]
    \centering
    \tiny
    \setlength{\tabcolsep}{2.2pt}
    \renewcommand{\arraystretch}{1.08}
    \caption{Experimental results on the Thames River Real and Thames River Simulated datasets. AP@0.5 (\%) is reported. Venues are omitted here for compactness; they are given in Table~\ref{tab:ship_lidar_results}, except LION (NeurIPS 2024). $F$ denotes effective FLOPs (G) averaged over validation samples; FPS is measured with batch size 1. Best AP results are highlighted in \textbf{bold}. ``--'' indicates that the category is unavailable.} 
    \label{tab:thames_full_metrics_aligned}
    \resizebox{\linewidth}{!}{
    \begin{tabular}{l|cccccc|ccccccc}
        \toprule
        \multirow{2}{*}{Method}
            & \multicolumn{6}{c|}{Thames River Real}
            & \multicolumn{7}{c}{Thames River Simulated} \\
        \cmidrule(lr){2-7} \cmidrule(lr){8-14}
        & Small & Med. & Large & mAP (\%) & $F$ (G) & FPS (Hz)
        & Small & Med. & Large & Buoy & mAP (\%) & $F$ (G) & FPS (Hz) \\
        \midrule
        PointPillars \cite{lang2019pointpillars}
            & 47.32 & 51.47 & 46.60 & 48.46 & 27.12 & 56.32
            & 62.00 & 69.00 & 77.80 & 51.80 & 65.20 & 41.22 & 69.93 \\
        PV-RCNN \cite{shi2020pv}
            & 38.10 & 82.70 & 82.40 & 67.74 & 510.09  & 7.18
            & 76.70 & 78.10 & 78.00 & 75.10 & 76.90 & 662.11 & 12.12 \\
        Part-A2 \cite{shi2019part}
            & 44.65 & 86.50 & 77.30 & 69.48 & 422.31 & 13.14
            & -- & -- & -- & -- & -- & -- & -- \\
        Voxel R-CNN \cite{deng2021voxel}
            & 40.14 & 84.63 & 81.62 & 68.80 & 12.17  & 24.87
            & 75.70 & 74.50 & 77.10 & 72.10 & 74.85 & 8.52 & 35.54 \\
        VoxelNeXt \cite{chen2023voxelnext}
            & 54.35 & 86.96 & 73.29 & 71.20 & 32.01 & 23.54
            & 87.16 & 82.56 & 86.50 & 63.16 & 79.84 & 22.40 & 22.09 \\
        DSVT \cite{wang2023dsvt}
            & 74.12 & 91.43 & 79.52 & 81.69 & 337.22 & 21.85
            & 85.50 & 76.10 & 78.70 & 77.60 & 79.47 & 374.12 & 14.28 \\
        LION \cite{liu2024lion}
            & -- & -- & -- & -- & -- & --
            & 90.60 & 75.4 & 78.30 & \textbf{85.1} & 82.4 & 1087.67 &28.12 \\
        FSHNet \cite{liu2025fshnet}
            & 42.21 & 75.94 & 69.93 & 62.69 & 46.30 & 16.27
            & 81.43 & 75.46 & 79.38 & 68.51 & 76.19 & 32.41 & 15.37 \\
        Fade3D \cite{ye2025fade3d}
            & 51.05 & 85.20 & 81.75 & 72.67 & 41.22 & 51.53
            & 84.92 & \textbf{83.14} & 88.20 & 64.10 & 80.09 & 44.91 & 52.31 \\
        GBlobs \cite{malic2025gblobs}
            & 67.09 & 86.04 & \textbf{96.83} & 83.32 & 31.01 & 22.74
            & \textbf{91.14} & 83.12 & 86.55 & 71.08 & 82.98 & 22.40 & 21.34 \\
        AutoReg3D \cite{huang2026feasibility}
            & 2.50 & 71.84 & 86.11 & 53.48 & 108.21 & 5.24
            & 14.83 & 33.79 & 53.04 & 36.50 & 34.54 & 120.31 & 1.88 \\
        \midrule
        \textbf{KSG-Net (Ours)}
            & \textbf{80.53} & \textbf{94.98} & 88.14 & \textbf{87.89} & 35.10 & 20.09
            & 90.72 & 82.63 & \textbf{89.31} & 75.11 & \textbf{84.44} & 24.56 & 19.85 \\
        \bottomrule
    \end{tabular}}
    \vspace{-6mm}
\end{table*}

\begin{figure*}[!t]
    \centering
    \includegraphics[width=0.85\textwidth]{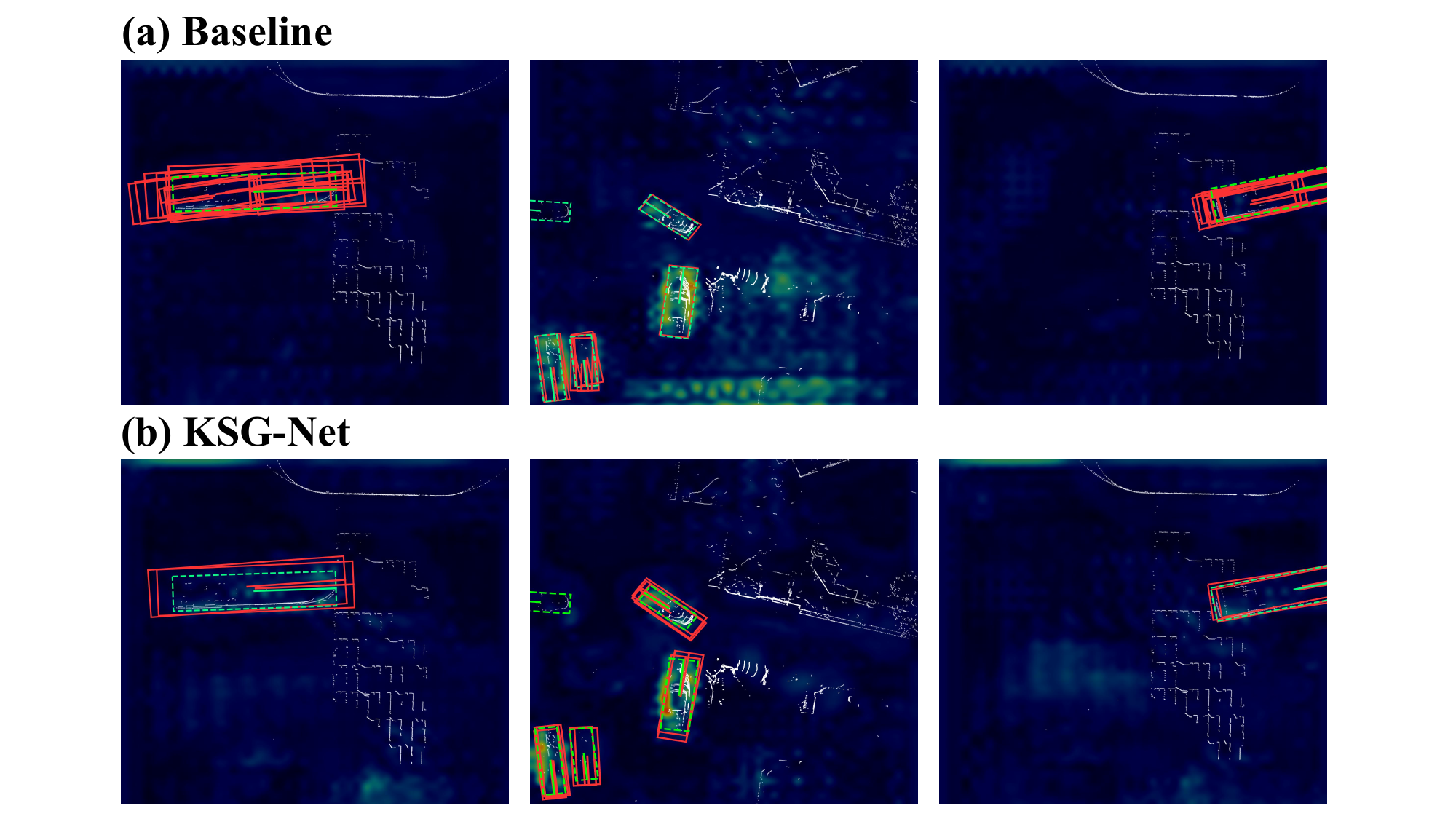}
    \vspace{-3mm}
    \caption{Feature visualization on the Thames River Real. 
    The heatmap shows the spatial activation patterns of KSG-Net in real maritime scenes. Stronger responses concentrate on vessel-related sparse regions and remain coherent along extended hulls, indicating that the full model preserves both local sparse cues and scene-level vessel structure.}
    \label{fig:real_heatmap}
    \vspace{-6mm}
\end{figure*}

As shown in Table~\ref{tab:thames_full_metrics_aligned}, KSG-Net achieves the best overall mAP on both the Thames River Real and Thames River Simulated datasets. On the Thames River Real Dataset, KSG-Net obtains an mAP of $87.89\%$, outperforming GBlobs~\cite{malic2025gblobs} by $4.57$ \% and DSVT~\cite{wang2023dsvt} by $6.20$ \%. On the Thames River Simulated Dataset, KSG-Net achieves an mAP of $84.44\%$, exceeding GBlobs ($82.98\%$), LION~\cite{liu2024lion} ($82.4\%$), and Fade3D~\cite{ye2025fade3d} ($80.09\%$). AutoReg3D~\cite{huang2026feasibility} remains clearly behind on both splits. These results indicate that key-sparse aggregation and global context injection generalize across real and simulated maritime scenes.
As shown in Fig.~\ref{fig:real_heatmap}, KSG-Net produces focused responses around sparse small-vessel regions and coherent activations along extended hulls. This qualitative pattern is consistent with Table~\ref{tab:thames_full_metrics_aligned}, where KSG-Net attains the highest Small and Medium AP on the Thames River Real Dataset together with the best overall mAP.
A closer category-wise comparison further highlights the effectiveness of KSG-Net for multi-scale vessel detection. On the Thames River Real Dataset, KSG-Net attains the highest Small and Medium AP ($80.53\%$ and $94.98\%$) and the highest overall mAP ($87.89\%$). Relative to DSVT, Small-vessel AP increases by $6.41$ \%. GBlobs records a higher Large-vessel AP ($96.83\%$ vs. $88.14\%$), while KSG-Net remains stronger in the sparse small-vessel regime and in overall mAP, which matches the role of KSMA.

\begin{figure*}[!t]
    \centering
    \includegraphics[width=\linewidth]{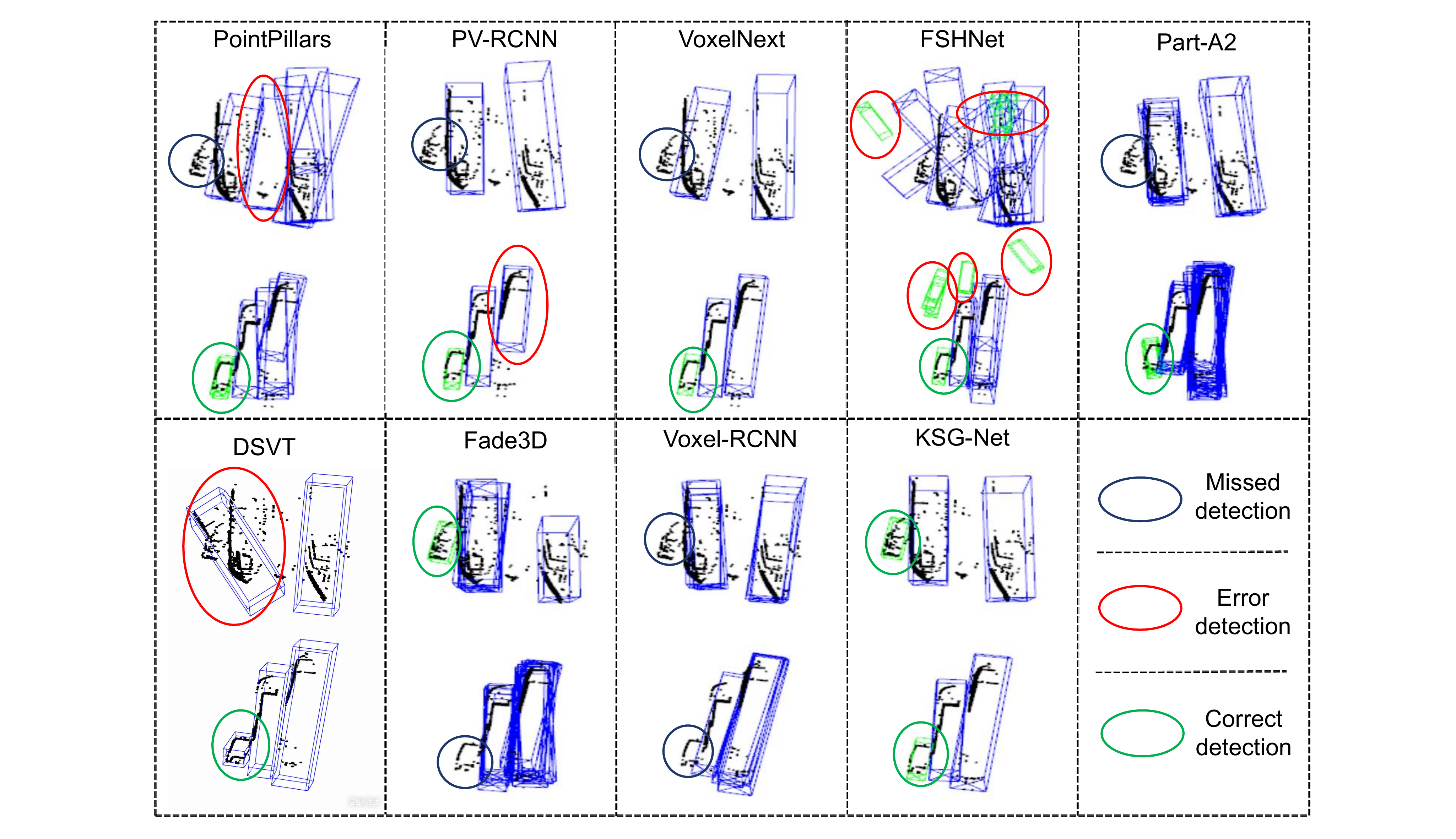}
    \vspace{-5mm}
    \caption{Qualitative comparison on Thames River Real. 
    Blue boxes denote predicted 3D bounding boxes. 
    The dark-blue, red, and green ellipses indicate missed detections, error detections, and correct detections, respectively. 
    Compared with existing methods, KSG-Net produces more accurate and stable predictions for vessels with different scales and reduces missed or erroneous detections in complex real maritime scenes.}
    \label{fig:real_visual_comparison}
    \vspace{-2mm}
\end{figure*}

As shown in Fig.~\ref{fig:real_visual_comparison}, the qualitative results on the Thames River Real Dataset further demonstrate the advantage of KSG-Net in complex maritime scenes. Existing methods tend to suffer from missed detections or inaccurate localization, especially under sparse point responses or partial observations. In contrast, KSG-Net produces more complete and stable 3D bounding boxes, particularly for small and medium vessels that are easily missed. This visual comparison is consistent with the Real Small and Medium AP and the overall mAP in Table~\ref{tab:thames_full_metrics_aligned}.

On the Thames River Simulated Dataset, KSG-Net maintains strong and balanced performance across vessel scales, achieving the best Large-vessel AP ($89.31\%$) and the best overall mAP ($84.44\%$). GBlobs obtains a slightly higher Small-vessel AP ($91.14\%$ vs. $90.72\%$). Although LION obtains a higher AP on the Buoy category, KSG-Net shows stronger overall detection capability on the vessel-related categories. This suggests that the proposed key-sparse aggregation and global context injection are particularly beneficial for maritime vessel detection, where targets often exhibit large-scale variations and irregular point distributions.

\subsection{Ablation Study}
To evaluate the individual contributions and complementarity of KSMA and GCA, we conduct ablation studies on both Thames River datasets (Table~\ref{tab: ablation_real_detailed_ksma_gca}). Here, \emph{Baseline} denotes our VoxelNeXt2D implementation with the IoU branch retained and both KSMA and GCA disabled; it is therefore different from the original VoxelNeXt configuration reported in Table~\ref{tab:thames_full_metrics_aligned}. The full framework improves mAP from $78.74\%$ to $87.89\%$ ($+9.15$\%) on Thames Real and from $77.13\%$ to $84.44\%$ ($+7.31$\%) on Thames Sim.

A closer look reveals complementary effects across categories. On Thames Real, KSMA improves Small-vessel AP by $17.37$\%; on Thames Sim, it improves Buoy AP by $16.32$\%. GCA alone raises overall mAP by $2.18$\% and $2.36$\% on the Real and Simulated datasets. Combining both modules yields the best mAP on both splits. On Real Large vessels, the baseline remains highest ($94.39\%$), whereas the full model trades part of this category for substantial Small/Medium and mAP gains. This indicates that KSMA and GCA work best as a joint multi-scale design: key-sparse enhancement recovers weak targets, while global context improves scene-level detection, with the largest benefit appearing in overall mAP.

\begin{table*}[!t] 
        \centering
        \tiny
        \caption{Ablation results on Thames Real and Thames Simulated datasets. The best results in each column are highlighted in \textbf{bold}.}
        \label{tab: ablation_real_detailed_ksma_gca}
        \setlength{\tabcolsep}{3.5pt} 
        \resizebox{\linewidth}{!}{
        \begin{tabular}{l|ccc|c|cccc|c}
            \toprule
            \multirow{3}{*}{Method} &  \multicolumn{4}{c|}{Thames Real Dataset} & \multicolumn{5}{c}{Thames Simulated Dataset} \\
            \cmidrule(lr){2-10}
             & \multicolumn{3}{c|}{AP@0.5 (\%)} & \textbf{mAP} & \multicolumn{4}{c|}{AP@0.5 (\%)} & \textbf{mAP}\\
            \cmidrule(lr){2-4}
            \cmidrule(lr){6-9}
             & Small & Med. & Large & (\%) & Small & Med. & Large & Buoy & (\%)\\
            \midrule
            Baseline & 57.86 & 83.98 & \textbf{94.39} & 78.74 & 86.94 & 81.39 & 87.18 & 53.03 & 77.13 \\
            \midrule
            Baseline + KSMA & 75.23 & 91.31 & 84.62 & 83.72 & 87.62 & 82.07 & 88.32 & 69.35 & 81.84 \\
            Baseline + GCA  & 65.32 & 83.26 & 94.18 & 80.92 & 86.33 & \textbf{82.73} & 87.42 & 61.48 & 79.49 \\
            \midrule
            \textbf{Ours} & \textbf{80.53} & \textbf{94.98} & 88.14 & \textbf{87.89} & \textbf{90.72} & 82.63 & \textbf{89.31} & \textbf{75.11} & \textbf{84.44} \\
            \bottomrule
        \end{tabular}} 
        \vspace{-4mm} 
\end{table*}
\subsection{Hyperparameter Analysis}
\begin{figure*}[!t]
    \centering
    \includegraphics[width=\textwidth]{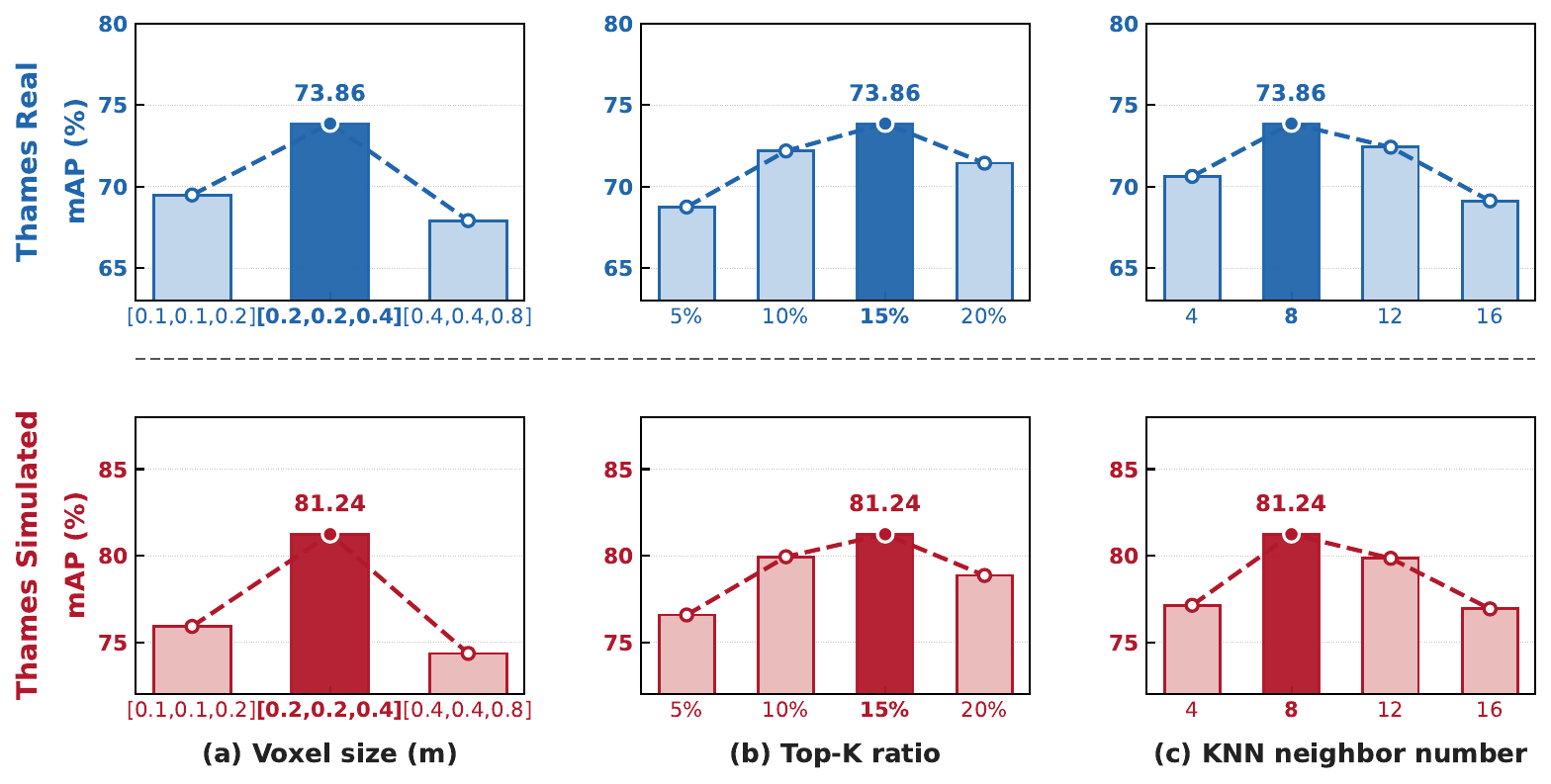}
    \vspace{-8mm}
    \caption{Hyperparameter analysis of the proposed method on the Thames River Real and Thames River Simulated datasets. 
    We investigate three key hyperparameters: 
    (a) voxel size, 
    (b) Top-K ratio for key-voxel selection, and 
    (c) KNN neighbor number. 
    The best performance on both datasets is achieved with a voxel size of $[0.2,0.2,0.4]$, a Top-K ratio of $15\%$, and a KNN neighbor number of $8$.}
    \label{fig:hyperparameter}
    \vspace{-6mm}
\end{figure*}
To further analyze the influence of key hyperparameters in the proposed method, we conduct hyperparameter experiments on the Thames River Real and Thames River Simulated datasets. As shown in Fig.~\ref{fig:hyperparameter}, we investigate three important hyperparameters, including the voxel size, the Top-K ratio for key-voxel selection, and the KNN neighbor number. Overall, these hyperparameters show consistent trends on both datasets, indicating that the proposed framework has relatively stable parameter sensitivity.

\noindent\textbf{Voxel size.}
Fig.~\ref{fig:hyperparameter}(a) shows the effect of voxel size. When the voxel size increases from $[0.1,0.1,0.2]$ to $[0.2,0.2,0.4]$, the performance improves noticeably on both datasets, while a further increase to $[0.4,0.4,0.8]$ leads to performance degradation. This indicates that overly fine voxels may produce fragmented sparse representations, whereas overly coarse voxels tend to lose local geometric details. Therefore, a moderate voxel size provides a better balance between structural completeness and local detail preservation.

\noindent\textbf{Top-K ratio.}
Fig.~\ref{fig:hyperparameter}(b) presents the influence of the Top-K ratio. As the ratio increases from $5\%$ to $15\%$, the detection performance consistently improves, while a slight decrease is observed at $20\%$. This suggests that selecting too few key voxels may miss informative sparse responses, whereas selecting too many may introduce redundant or noisy locations and weaken the selectivity of KSMA.

\noindent\textbf{KNN neighbor number.}
Fig.~\ref{fig:hyperparameter}(c) further analyzes the effect of the KNN neighbor number. The best performance on both datasets is achieved when the neighbor number is set to $8$. A smaller number of neighbors limits local context modeling, while an excessively large number may introduce irrelevant neighborhood information and reduce the precision of feature aggregation.

\section{Conclusion}
In this paper, we explore a practical and challenging problem of maritime 3D ship detection, which requires accurately detecting vessels of varying scales under sparse point observations and severe background interference. To tackle the above issue, we propose KSG-Net, a Key-Sparse and Global-Context learning network tailored for maritime 3D ship detection. The core idea is to jointly enhance local discriminative representations for small vessels and global structural awareness for large vessels within a unified fully sparse detection framework. To this end, two complementary modules are designed: a Key Sparse Multi-scale Aggregation (KSMA) module to enhance the representation of small and sparse vessels through informative key-voxel selection and cross-scale neighborhood aggregation; and a Global Context Aggregation (GCA) module to capture long-range geometric dependencies for large vessels through scene-level global context modeling with gated residual interactions. Extensive experiments on the Thames River vessel dataset and simulated datasets demonstrate that KSG-Net consistently outperforms existing methods in multi-scale vessel detection and exhibits strong robustness in complex maritime environments, validating the effectiveness of the proposed KSMA and GCA modules.

\bibliographystyle{splncs04}
\bibliography{references}

\end{document}